\documentclass[sigconf]{acmart}
\usepackage{times}
\usepackage{soul}
\usepackage{url}
\usepackage[utf8]{inputenc}
\usepackage{graphicx}
\usepackage{makecell}
\usepackage{amsmath}
\usepackage{amsthm}
\usepackage{booktabs}
\usepackage[switch]{lineno}
\usepackage{epsfig}
\usepackage{multirow}
\usepackage{tabularx}
\usepackage{dsfont}
\usepackage{url}
\usepackage{color}
\usepackage{xcolor}
\usepackage{colortbl}
\usepackage{diagbox}
\usepackage[linesnumbered,titlenumbered,ruled,vlined,commentsnumbered,noend]{algorithm2e}
\SetKwComment{Comment}{$\triangleright$\ }{}
\usepackage[export]{adjustbox}
\usepackage{comment}
\usepackage{pifont}
\usepackage{indentfirst}

\usepackage[T1]{fontenc}    
\usepackage{enumitem}
\usepackage{amsfonts}       
\usepackage{nicefrac}       
\usepackage{microtype}      
\usepackage[rightcaption]{sidecap}
\usepackage{wrapfig}
\usepackage{stfloats}

\definecolor{dg}{rgb}{0,0.694,0.298}
\definecolor{purple}{rgb}{0.4,0.176,0.569}
\definecolor{Gray}{gray}{0.6}
\definecolor{royalblue}{RGB}{65,105,225}

\usepackage[capitalize]{cleveref}
\crefname{section}{Sec.}{Secs.}
\Crefname{section}{Section}{Sections}
\Crefname{table}{Table}{Tables}
\crefname{table}{Tab.}{Tabs.}

\makeatletter
\DeclareRobustCommand\onedot{\futurelet\@let@token\@onedot}
\def\@onedot{\ifx\@let@token.\else.\null\fi\xspace}
\def\eg{\emph{e.g}\onedot} 
\def\ie{\emph{i.e}\onedot} 
 
\def\etc{\emph{etc}\onedot} 
 
\def\etal{\emph{et al}\onedot}
\makeatother

\begin{document}
\title{Architecture-agnostic Iterative Black-box Certified Defense against Adversarial Patches}


\author{
Di Yang$^1$, \ Yihao Huang$^2$, \ Qing Guo$^{3}$, \ Felix Juefei-Xu$^4$, \ Ming Hu$^2$, \ Yang Liu$^{2}$, \ Geguang Pu$^1$}
\affiliation{\institution{
$^1$East China Normal University \country{China} 
$^2$Nanyang Technological University \country{Singapore}\\
$^3$Centre for Frontier AI Research (CFAR), A*STAR \country{Singapore}
$^4$New York University \country{USA}}}


\renewcommand{\shortauthors}{Di Yang, et al.}

\begin{abstract}
The adversarial patch attack aims to fool image classifiers within a bounded, contiguous region of arbitrary changes, posing a real threat to computer vision systems (\eg, autonomous driving, content moderation, biometric authentication, medical imaging) in the physical world. 
To address this problem in a trustworthy way, proposals have been made for \textit{certified} patch defenses that ensure the robustness of classification models and prevent future patch attacks from breaching the defense.
State-of-the-art certified defenses can be compatible with any model architecture, as well as achieve high clean and certified accuracy.
Although the methods are adaptive to arbitrary patch positions, they inevitably need to access the size of the adversarial patch, which is unreasonable and impractical in real-world attack scenarios. 
To improve the feasibility of the architecture-agnostic certified defense in a black-box setting (\ie position and size of the patch are both unknown), we propose a novel two-stage Iterative Black-box Certified Defense method, termed \textbf{IBCD}.
In the first stage, it estimates the patch size in a search-based manner by evaluating the size relationship between the patch and mask with pixel masking. In the second stage, the accuracy results are calculated by the existing white-box certified defense methods with the estimated patch size. 
The experiments conducted on two popular model architectures and two datasets verify the effectiveness and efficiency of IBCD.
\end{abstract}




\maketitle
\section{Introduction}
\begin{figure}[t]
    \centering
    \includegraphics[width=\linewidth]{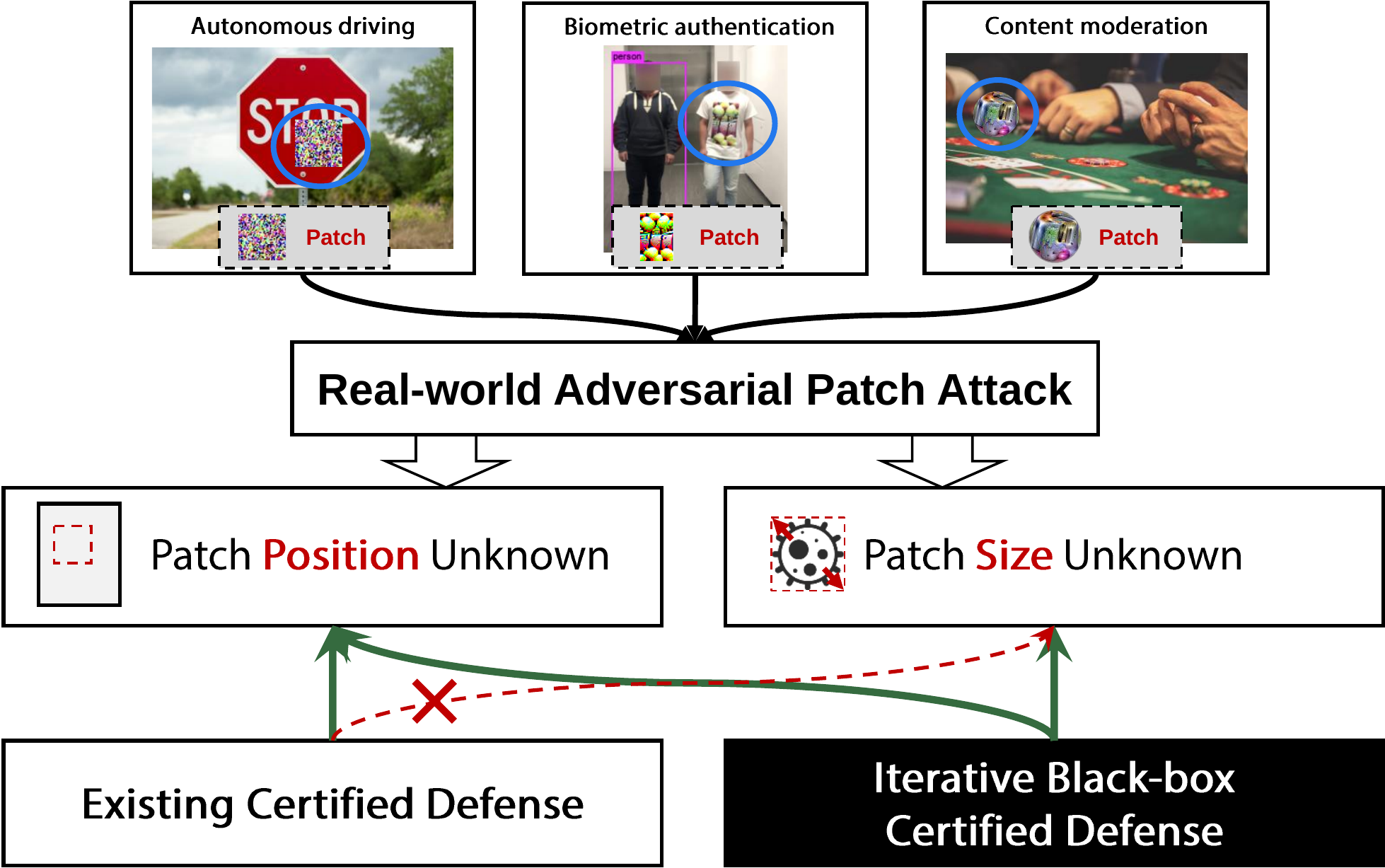}
    \caption{In real-world scenarios, adversarial patch attacks may exist in autonomous driving \cite{brown2017adversarial}, content moderation \cite{liu2018dpatch}, biometric authentication \cite{xu2020adversarial}, \etc. In these situations, the patch size and patch position are unknown. Existing certified defenses are white-box methods that can only handle the ``patch position unknown'' problem. Compared with them, the proposed iterative black-box certified defense method can further address the ``patch size unknown'' problem.}
    \label{fig:IBCD_compare_white-box}
\end{figure}

Adversarial patch attacks, as a real threat that can be implemented against multimedia applications with carriers such as stickers and paints in the ``physical world'' (\eg, adding an adversarial sticker on a traffic sign to fool self-driving cars \cite{brown2017adversarial,karmon2018lavan}, adding an adversarial patch on images for evading content moderation of social media platforms \cite{liu2018dpatch}, adding an adversarial pattern on the clothes to fool biometric authentication \cite{xu2020adversarial}, \etc), has become a topic of great interest in recent years. It allows the attackers to modify a bounded continuous region (usually defined as a square patch) of any position in an image. Obviously, unrestrained attack strength and position bring a severe challenge to adversarial defense methods, leading to the total failure of traditional empirical adversarial defense methods \cite{naseer2019local,hayes2018visible} when countering with adaptive white-box patch attacks \cite{tramer2020adaptive}.

In order to realize provable security, \textit{certified} patch defense \cite{chiang2020certified} has been proposed to guarantee the robustness of models against any adversarial patches (even a patch of which any pixel can misclassify the model) without empirical evaluation. Early certified defense methods \cite{lin2020certified,metzen2021efficient,xiang2021patchguard,zhang2020clipped} heavily depend on specific model architectures with small receptive fields, which significantly reduce the clean accuracy result. Recent works have implemented architecture-agnostic certifiably robust image classification as well as achieving remarkable performance by derandomized smoothing \cite{salman2022certified,levine2020randomized,chen2022towards} or pixel masking \cite{xiang2022patchcleanser}. However, state-of-the-art works inevitably need to access the size of the adversarial patch, which is unreasonable and impractical in real attack scenarios. For the sake of distinction, we call these defense methods white-box certified defense since they need to know the patch size.  

To design the architecture-agnostic certified defense in a black-box setting (\ie, patch position and size are unknown), we propose a novel two-stage Iterative Black-box Certified Defense method, termed \textbf{IBCD}. In the first stage, it estimates the patch size in a search-based manner with three main components: search operation, satisfiability check, and search space reduction. To be specific, in each iteration, the method first applies the search operation (\eg, pixel masking) to collect the prediction results of the input image and then performs a satisfiability check to judge the relationship between mask and patch. If the mask is bigger than the patch, then perform search space reduction and start the next iteration. The search procedure stops when the mask is smaller than the patch and the patch size strictly falls between the mask size in the current iteration and the previous one. In the second stage of IBCD, the estimated patch size can support the calculation of the clean and certified accuracy with existing certified defense methods. 

In summary, our work has the following contributions:
\begin{itemize}
\item To the best of our knowledge, we are the first to propose architecture-agnostic certified defense in a black-box setting (\ie, the position and size of the patch are unknown), which promotes the practicality of certified defense for protecting multimedia applications in the physical world.
\item We design a search-based algorithm to efficiently estimate the size of the adversarial patch. We also propose a sliding space optimization strategy to accelerate the search for better efficiency. 
\item The experiment conducted on two popular datasets (\ie, ImageNet and CIFAR10) with two representative model architectures (\ie, ResNet and ViT) shows the efficiency and usability of IBCD.
\end{itemize}

\section{Related Work}
Chiang \cite{chiang2020certified} proposed the first certified defense against adversarial patch attacks. However, it suffers from extremely expensive model training and is not applicable to high-resolution images. Early certified defenses are strongly model-dependent \cite{zhang2020clipped,xiang2021patchguard,metzen2021efficient}, which lack generality. State-of-the-art certified patch defense methods against adversarial patches are pursuing architecture-agnostic and are available on high-resolution images, falling into two main categories: derandomized smoothing and pixel masking.

\subsection{Derandomized Smoothing}
Derandomized smoothing fed small image regions (also called image ablations) to a classification model and performed the majority voting for the final prediction.
For adversarial patch, Levine \etal{} propose Derandomized Smoothing (DS) \cite{levine2020randomized}, which trains a base classifier by smoothed images and the final classification is based on majority votes. This method significantly improves the accuracy of certified patch defense on ImageNet but inference computation is expensive. ECViT \cite{chen2022towards} propose a progressive smoothed image modeling task to train Vision Transformer, which can capture the more discriminable local context of an image while preserving the global semantic information. The Smoothed ViT \cite{salman2022certified} shows that Derandomized smoothing combined with Vision Transformers can significantly improve certified patch robustness that is also more computationally efficient and does not incur a substantial drop in standard accuracy.

\subsection{Pixel Masking}
Pixel masking recovers the correct prediction with high
probability if all the patch regions are masked.
PatchCleanser \cite{xiang2022patchcleanser} uses two rounds of pixel masking to eliminate the effects of patch attacks. PatchCleanser masks the image before the input layer which can be compatible with models of any architecture and achieved state-of-the-art certified robust accuracy.

\section{Preliminaries and Motivation}\label{sec:preliminaries}

\subsection{Preliminaries}
\paragraph{Adversarial patch attack.} It aims to fool an image classifier within a bounded contiguous region of arbitrary changes anywhere on the image. To be specific, given a classification model $\mathcal{F}(\cdot)$ and an input image $\mathbf{I} \in [0,1]^{H \times W \times 3}$ with ground truth label $y$, the purpose of the adversarial patch attack is to generate an adversarial example $\mathbf{\hat{I}} \in \mathcal{A}(\mathbf{I},\delta)$ satisfies $\mathcal{F}(\mathbf{\hat{I}}) \ne y$, where $\mathcal{A}$ is the threat model and $\delta$ is the adversarial patch. In general, the patch $\delta \in [0,1]^{v \times v \times 3}$ is a square block and $v$ is recorded briefly as the size of the patch. 

\paragraph{Certified patch defense.} It aims to construct a defended model $\mathcal{D}$ that can always give a correct prediction for adversarial examples generated by any attack within the threat model $\mathcal{A}$, \ie, $\forall \mathbf{\hat{I} \in \mathcal{A}(\mathbf{I},\delta)}$, $\mathcal{D}(\mathbf{I})=\mathcal{D}(\mathbf{\hat{I}})=y$. Note that threat model $\mathcal{A}$ could fully access defense method, model parameters, \etc. The certification calculates a \textit{provable lower bound} on the model robustness against adaptive white-box attacks. 

\subsection{Problem of State-of-the-art White-box Certified Defenses}
For both \textit{derandomized smoothing} and \textit{pixel masking}, the certified patch defenses have an obvious defect, \ie, they can not cope with the situation that the patch size is unknown. For \textit{pixel masking-based} certified defense, they claim that ``the defender has a conservative estimation of the patch size'' \cite{xiang2022patchcleanser}. For \textit{derandomized smoothing-based} certified defense, the patch size is an important factor used in the formula that guarantees the certified robustness (See Eq.~\eqref{eq:Derandomized_Smoothing_ceritified},\eqref{eq:Derandomized_Smoothing_block_delta},\eqref{eq:Derandomized_Smoothing_band_delta}). They can not calculate the certified accuracy without patch size. However, it is unrealistic and unreasonable for the defender to accurately know the size of the patch in a real offensive and defensive environment. Therefore, designing a black-box (both patch position and patch size are unknown) certified patch defense method is urgent and of great practical significance. 

\subsection{Motivation and Premise}
Patch size estimation is the main challenge in black-box certified defense. Optimization-based methods seem not robust enough to handle the adversarial patch attack that can arbitrarily modify the patch position and size in an unpredictable manner. Because the attackers can easily design a deliberate adversarial attack with a comprehensive understanding of a specific defense method. Therefore, we consider the commonly used search-based algorithm as it has shown excellent attack performance in other black-box tasks such as black-box adversarial attacks and black-box testing.

\begin{figure}
    \centering
    \includegraphics[width=0.9\linewidth]{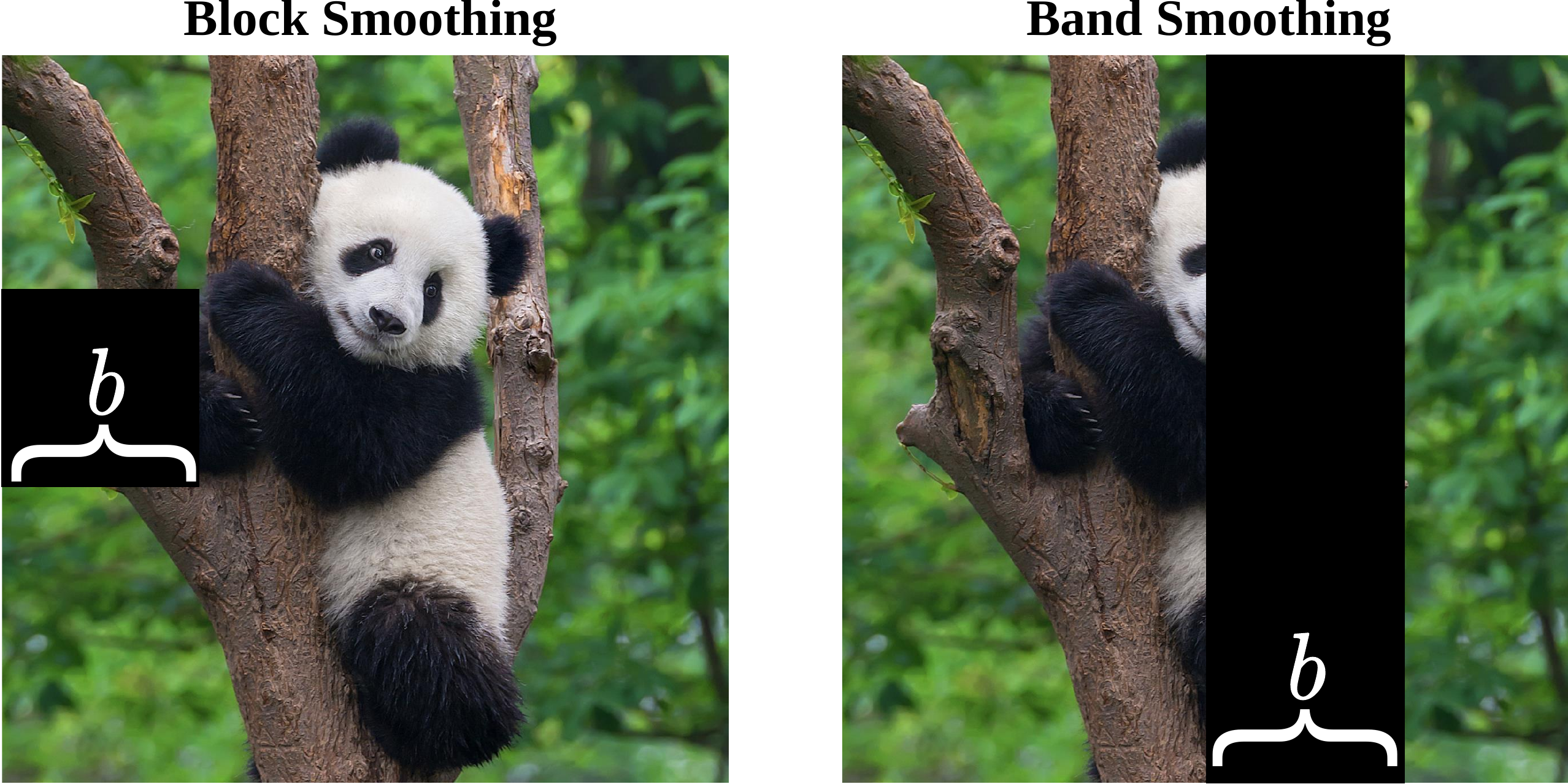}
    \caption{Two types of Derandomsized smoothing.}
    \label{fig:derandomsized_smoothing}
\end{figure}

It's worth noting that we design the black-box certified defense algorithm under the assumption that \textit{the size of the adversarial patch is not larger than a quarter of the image}. The assumption is necessary according to three reasons. \ding{182} For commonly used datasets (\eg, CIFAR10, MNIST, ImageNet), the objects in some of the images are smaller than a quarter of the image size. It is unrealistic to expect a classifier to accurately predict images in which the object is completely occluded by a patch of extremely large size. \ding{183} The setting of patch size limits the search space, which is helpful for confirming a reasonable initial state of the search algorithm. \ding{184} The size is approximate to the capability boundary of derandomized smoothing-based certified defense and pixel masking-based certified defense. 
To be specific, for \textbf{pixel masking} (\ie, PatchCleanser), on $224 \times 224$ images of ImageNet, its certified accuracy against the patch of size $112 \times 112$ is only 20.5\% with ViT, which means the patch larger than a quarter of the image can be hardly certified defended by PatchCleanser. 
The \textbf{derandomized smoothing} method aims to retain a part continuous region in the image while removing other parts of the image. There are two types of smoothing: block smoothing and band smoothing. Block/Band smoothing means removing the entire image except for a square-block/band region, where the width of the block/band is $b$. In Fig.~\ref{fig:derandomsized_smoothing} shows the image ablations of block smoothing and band smoothing, called block ablation and band ablation respectively. A block/band ablation can start at any position and wrap around the image, we set $K$ as the number of possible block/band ablations. It is approximated that $K= H \times W$ for block smoothing and $K = W$ for band smoothing. For each class $y$, 
\begin{align}
n_{y}(\mathbf{I}) = \sum_{k = 1}^{K} \mathcal{Q}(\mathcal{F}(\mathrm{Abl}(\mathbf{I}, b, k)) = y),
\label{eq:Derandomized_Smoothing_vote}
\end{align}
denotes the number of ablations that were classified as class $c$, where $\mathrm{Abl}(\mathbf{I}, b, k)$ is the $k$th image ablation and $\mathcal{Q}(\cdot)$ is 1 if the condition is true else returns 0.

With respect to the theory of Derandomized Smoothing \cite{levine2020randomized}, an image is certified robust if and only if the statistics of the highest class $y$ (\ie, $n_{y}(\mathbf{I})$) is a large margin bigger than the second highest class $y'$ (\ie, $n_{y'}(\mathbf{I})$), formulated as
\begin{align}
n_{y}(\mathbf{I})\ge \max _{y^{\prime} \neq y} n_{y^{\prime}}(\mathbf{I})+2\Delta,
\label{eq:Derandomized_Smoothing_ceritified}
\end{align}
where $\Delta$ is the maximum number of intersections between image ablations and the adversarial patch.

There is also an implied restriction for $n_{y}(\mathbf{I})$ and $n_{y^{\prime}}(\mathbf{I})$, that is, the sum of them should not exceed the number of possible block/band ablations (\ie, $K$), formulated as 
\begin{align}
n_{y}(\mathbf{I}) + n_{y^{\prime}}(\mathbf{I}) \le K.
\label{eq:Derandomized_Smoothing_limit}
\end{align}
With Eq.~\eqref{eq:Derandomized_Smoothing_ceritified} and Eq.~\eqref{eq:Derandomized_Smoothing_limit}, we can directly derive Eq.~\eqref{eq:Derandomized_Smoothing_conclusion}, 
\begin{align}
\Delta  \leq \frac{K}{2}-n_{y^{\prime}}(\mathbf{I}).
\label{eq:Derandomized_Smoothing_conclusion}
\end{align}
%
Thus for block smoothing, 
\begin{equation}
\begin{aligned}
\Delta =  
(v+b-1)^2,
\end{aligned}
\label{eq:Derandomized_Smoothing_block_delta}
\end{equation}
\begin{equation}
\begin{aligned}
v & \le \sqrt{\frac{K}{2} - n_{y^{\prime}}(\mathbf{I})}-b+1 \le \sqrt{\frac{K}{2}} = \sqrt{\frac{H \times W}{2}},
\end{aligned}
\label{eq:Derandomized_Smoothing_block_upper}
\end{equation}
which means the region of the patch is less equal than half of the image.
For band smoothing,
\begin{equation}
\begin{aligned}
\Delta =  
(v+b-1),
\end{aligned}
\label{eq:Derandomized_Smoothing_band_delta}
\end{equation}
\begin{equation}
\begin{aligned}
v & \le \frac{K}{2} - n_{y^{\prime}}(\mathbf{I})-b + 1 \le \frac{K}{2} = \frac{W}{2},
\end{aligned}
\label{eq:Derandomized_Smoothing_band_upper}
\end{equation}
which means the patch region is less equal than a quarter of the image. Among the boundaries of band and block smoothing, we choose the lower one (\ie, one-quarter of the image) to include all the possible derandomized smoothing types.

\section{Iterative Black-box Certified Defense}
\subsection{Formulation and Overview}
The existing defended model $\mathcal{D}(\cdot)$ of architecture-agnostic white-box methods (\ie, derandomized smoothing and pixel masking) are actually $\mathcal{D}(\mathbf{\hat{I}},v)$, where $v$ is the patch size and is known by them. In order to against the situation that the size and position of the patch are both unknown, we construct the architecture-agnostic black-box certified defense as a two-stage procedure: (1) patch size estimation, (2) certified defense with estimated patch size. The defended model can be reformulated as $\mathcal{D}(\mathbf{\hat{I}},{\mathcal{E}(\mathbf{\hat{I}}}))$, where $\mathcal{E}(\cdot)$ means patch size estimation. With respect to the second stage, the existing derandomized smoothing methods or pixel masking methods are the feasible choices. Thus the following introduction mainly focuses on the design of patch size estimation (\ie, $\mathcal{E}(\cdot)$).

\begin{figure}[t]
    \centering
    \includegraphics[width=0.8\linewidth]{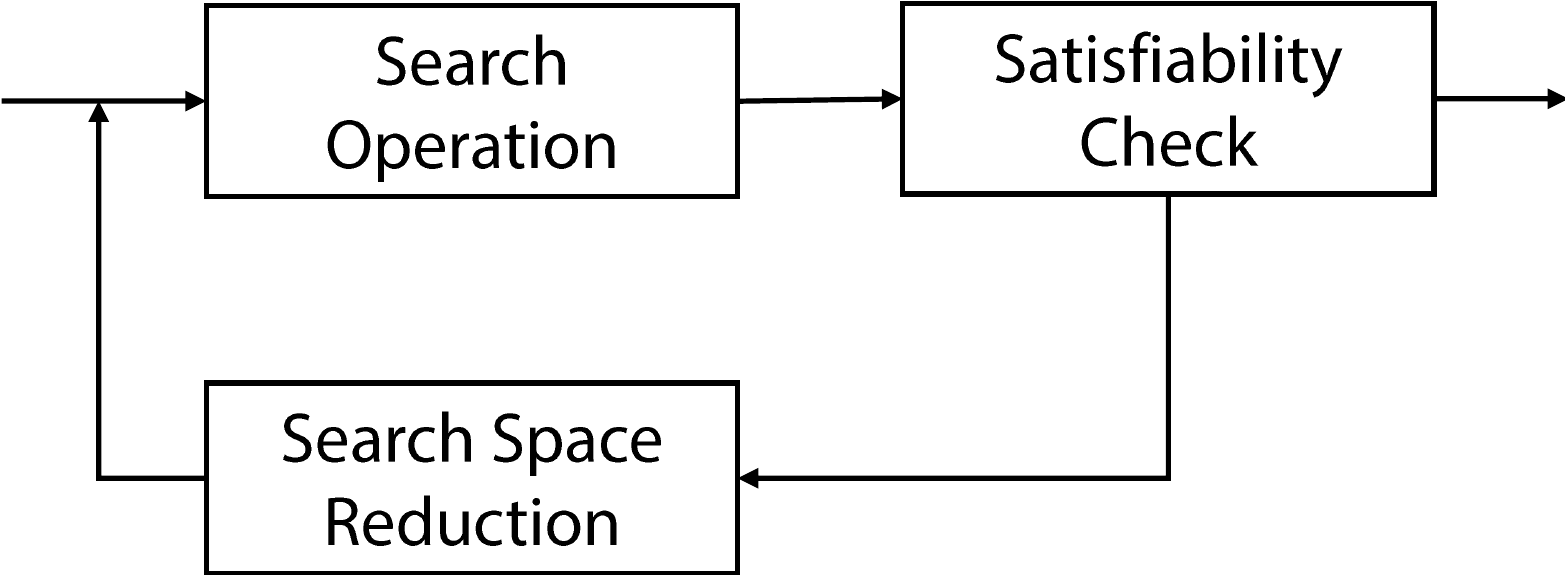}
    \caption{Overview of the patch size estimation framework.}
    \label{fig:patch_estimation_framework}
\end{figure}

In Fig.~\ref{fig:patch_estimation_framework}, the patch size estimation method is a search-based iterative scheme with three core components: search operation, satisfiability check, and search space reduction.
In each iteration, \textit{search operation} needs to evaluate the image and gives distinguishable results according to the size and position of the patch. \textit{Satisfiability check} analyzes the information collected from the \textit{search operation} and gives the conclusion that whether the search space needs to be reduced. If the search space can be further reduced, then perform \textit{search space reduction}, otherwise, stop the whole estimation procedure and the patch size can be estimated with the information provided in the last iteration and penult iteration. Please note that the design of \textit{search space reduction} is significant since it not only needs to reduce the search space steadily but also better to reduce the search space as more as possible while guaranteeing the patch size strictly falls within a certain range.

\subsection{Implementation Details}
\begin{figure*}[]
    \centering
    \includegraphics[width=0.9\linewidth]{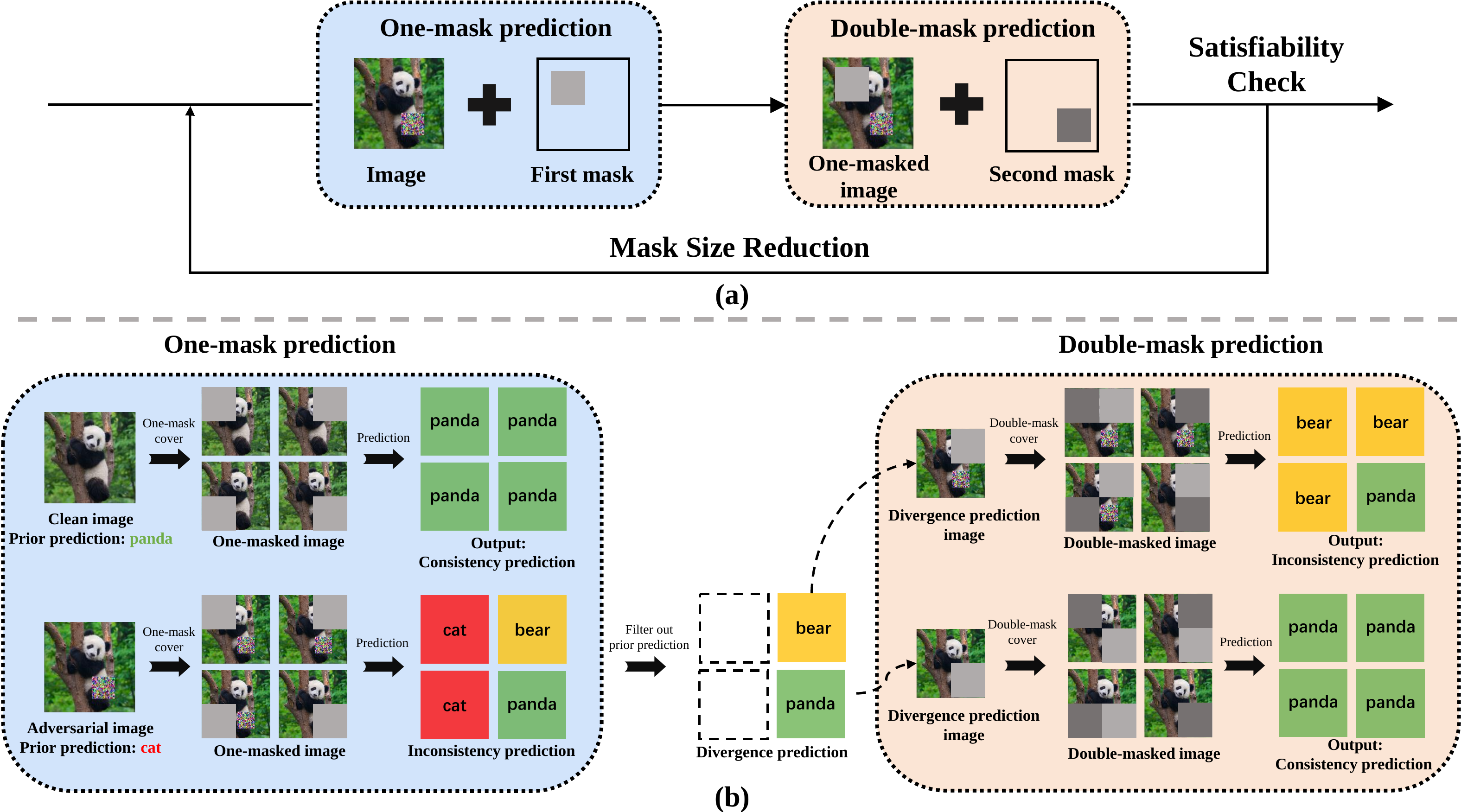}
    \caption{
    \textbf{Implementation details of patch size estimation.} We design a double masking-based search operation to realize the patch size estimation. In each iteration, the consistency predictions given by the search operation are collected into a set and the satisfiability check takes it as the input. If one of the predictions is ``consistent'', then perform mask size reduction and start the next iteration. If all the predictions are ``inconsistent'', then the mask is smaller than the patch and the search procedure is terminated.
    }
    \label{fig:Implementation_of_patch_size_estimation}
\end{figure*}

We will introduce the details of the core components (\ie, search operation, satisfiability check, and search space reduction), as an example (Fig.~\ref{fig:Implementation_of_patch_size_estimation}(a) and (b)) to show how to implement the framework mentioned in Fig.~\ref{fig:patch_estimation_framework}. Please note, in order to defend the most offensive patch, we assume the patch can successfully mislead the classifier with any of its pixels. 

\textbf{Search operation.}
We design \textit{search operation} with reference to the double-masking algorithm \cite{xiang2022patchcleanser} and modify it to satisfy our requirement. An overview of the search operation is shown in Fig.~\ref{fig:Implementation_of_patch_size_estimation}(b). It is a double-masking algorithm with two same masks of fixed size. Here each mask is a binary tensor $\mathbf{m} \in \{0,1\}^{H \times W}$, which has the same width and height as the image. The elements within the mask take values of 0, and others are 1. In the first round of masking (\ie, blue block of Fig.~\ref{fig:Implementation_of_patch_size_estimation}(b)), the input is a clean or adversarial image. We first put the image into the classifier to achieve a prior prediction (in Fig.~\ref{fig:Implementation_of_patch_size_estimation}, assume ``panda'' for the clean image and ``cat'' for the adversarial image). Then we use a mask to cover the image as a sliding window and puts the generated one-masked images into the classifier. If these one-masked images have a consistent prediction result and are the same as the prior prediction, then give an output $\mathcal{CP}=1$, which represents ``consistency prediction''. If the images have inconsistent prediction results, then the input image must be an adversarial image. Please note that we do not know the relationship between the size of the mask and the patch, thus we discuss the method under two different cases. \ding{182} \textbf{Mask equal or bigger than patch} In this case, there at least exists a one-masked image that covers the patch and is a correct label (Fig.~\ref{fig:Implementation_of_patch_size_estimation} shows this case). There may exist multiple wrong labels. To simplify, we assume there are two wrong labels ``cat'' and ``bear'', where ``cat'' is the same as the prior prediction. Please note that the operation towards more wrong labels is the same. To reduce computation, we then filter out the images that have the same prediction as prior (\ie, ``cat''). Because they still lead to inconsistency in double-mask prediction and the number of inconsistency predictions does not influence the result of the satisfiability check. Here it is confusing which label of ``bear'' and ``panda'' is the correct label, thus second-round masking is necessary. In second-round masking, the images of ``bear'' or ``panda'' will lead to different prediction results. For images of the label ``bear'', there must exist a double-masked image in which the adversarial patch is fully covered by the second mask and returns ``panda'', thus leading to inconsistency prediction. For images of the label ``panda'', since the first mask fully covers the patch, then the predictions of all the double-masked images should be the same. We can confirm that the labels in consistency prediction are the correct label (\ie, ``panda''). To summarize, in the case that the mask is equal to or bigger than the patch, the conclusion is certain and the correct label is known. \ding{183} \textbf{Mask smaller than patch} In this case, the mask can never cover the patch and all the predictions are wrong. If the predictions are all the same as the prior prediction, we can easily output an inconsistency prediction (\ie, $\mathcal{CP}=0$). If the predictions are different, there exists a problem: it is impossible to judge whether the labels other than the prior are correct or not in the current iteration because we actually  do not know the relationship between the size of the mask and patch. To solve this problem, we need information from other iterations and this is why we assume \textit{the size of the adversarial patch is not larger than a quarter of the image}. According to this assumption, we start the first iteration of the search procedure with the setting that the mask is equal to or larger than a quarter of the image. At the first iteration, we can ensure the mask is larger than the patch and obtain the correct label. The label could be used as a reference to confirm whether the predictions of one-masked images are correct or not, which solves the problem mentioned before. To summarize, in the case that the mask is smaller than the patch, the output must be inconsistency predictions (\ie, $\mathcal{CP}=0$) with the help of the information obtained in the first iteration. 

\textbf{Satisfiability check.} The output collection of \textit{search operation} is a set $\mathbf{SCP}=\{{\mathcal{CP}_1,\mathcal{CP}_2},...\}$. We can conclude that the mask is equal or bigger than the patch if and only if ${\exists}\mathcal{CP}\in \mathbf{SCP}, \mathcal{CP}=1$. In this situation, the state $\mathcal{SCP}_s$ is True. We can conclude that the mask is smaller than the patch if and only if ${\forall}\mathcal{CP}\in \mathbf{SCP}, \mathcal{CP}=0$. In this situation, the state $\mathcal{SCP}_s$ is False.

\textbf{Search space reduction.} Please note the search space is generally referred to the situations that need to be checked (\ie, applying the search operation with different mask sizes to see the result of the satisfiability check), not the mask sliding space on the image. For the double-masking method, the way of search space reduction is to decrease the mask size as the number of iterations increases.

\textbf{Details of the mask.}
In order to ensure that the mask can cover all the positions where the patch may injection when it is equal to or bigger than the patch, we follow the method of calculating the mask step size in PatchCleanser \cite{xiang2022patchcleanser}. As shown in Eq.~\eqref{eq:multi_scale_mask_set_certified}, with the mask size $\eta$, sliding stride $s$ and actual patch size $v$, the constraint is
\begin{align}
v \le \eta - s + 1.
\label{eq:multi_scale_mask_set_certified}
\end{align}
The pre-defined multi-scale mask set $\mathbf{M}$ contains mask subsets of different sizes $\{\eta_{1},\eta_{2},...\}$. For example, $\mathbf{M}[\eta_{1}]$ is a set that contains masks that have the same size (\ie, $\eta_{1}$) and different positions. The masks in set $\mathbf{M}[\eta_{1}]$ can fully cover the image.


\subsection{Sliding Space Optimization}
\begin{figure}[t]
    \centering
    \includegraphics[width=\linewidth]{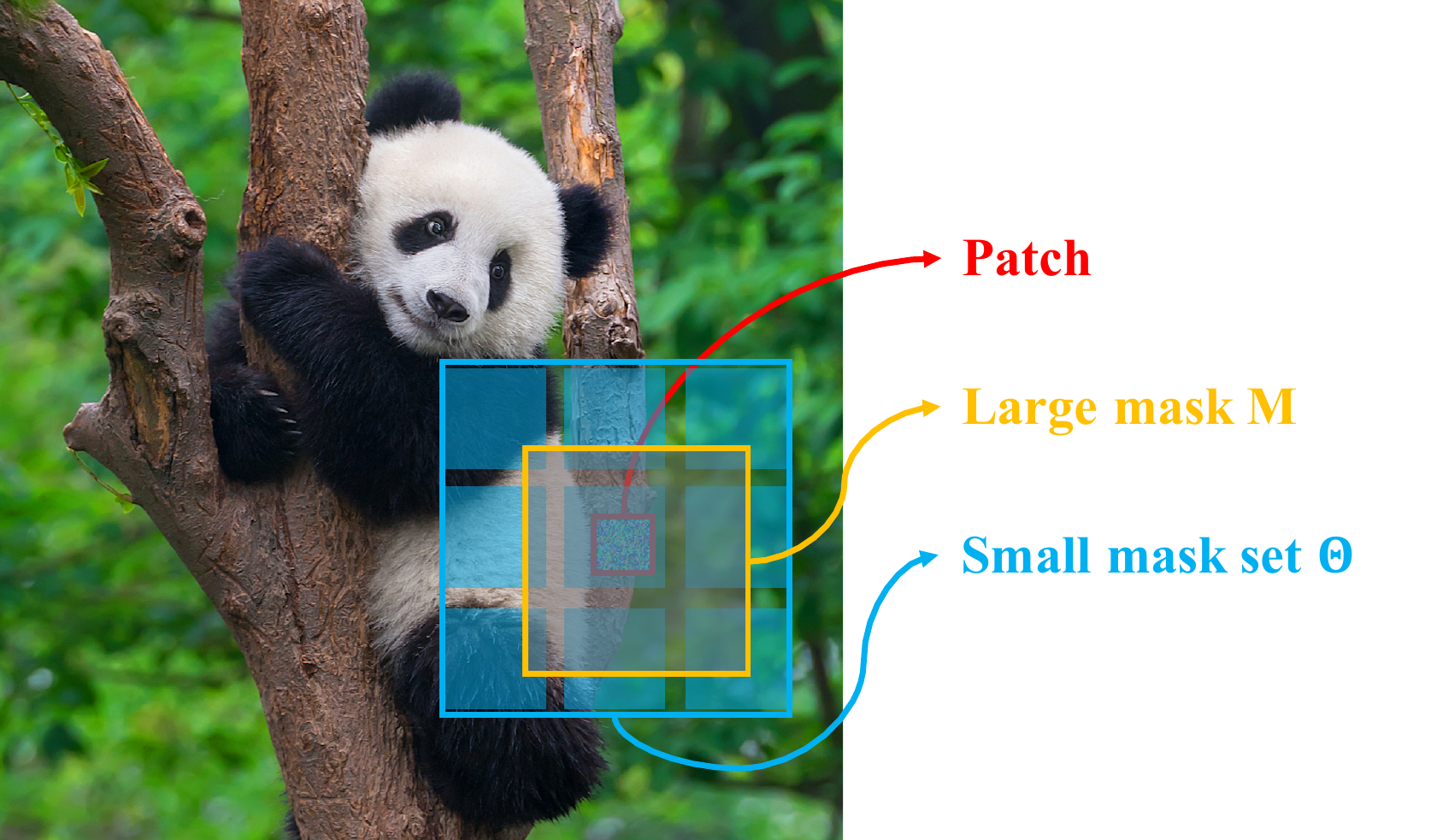}
    \caption{\textbf{Sliding space optimization.} In the case where the large mask (yellow box) covers the patch (red box), we only select the small masks (blue box) that intersect with the large mask to execute the search operation, rather than search on whole the image. This strategy can help to improve the efficiency of IBCD.}
    \label{fig:optimization}
\end{figure}
As a search-based algorithm, the efficiency of patch size estimation is an important metric. To further reduce the computational complexity and make it more practical, we propose a sliding space optimization method.

The high-level idea is that, once the mask successfully covers the patch, its position is the sliding space of the next iteration, which is much smaller than taking the whole image as the sliding space. As shown in Fig.~\ref{fig:optimization}, the red box and yellow box represent the position of the patch and the large mask that can completely cover the patch respectively. The blue box represents the set of all possible small mask positions that intersect with the large mask, which is the sliding space that needs to be explored. To be specific, we define 
\begin{align}
\mathcal{R}(\mathbf{m})  = \left[x_{1}^{\mathbf{m}}, y_{1}^{\mathbf{m}}, x_{2}^{\mathbf{m}}, y_{2}^{\mathbf{m}}\right]
\label{eq:mask_pisition}
\end{align}
as the region of mask $\mathbf{m}$, where $(x_{1}^{\mathbf{m}}, y_{1}^{\mathbf{m}})$ and $(x_{2}^{\mathbf{m}}, y_{2}^{\mathbf{m}})$ represent the position of the top left vertex and bottom right vertex of the mask. $\mathcal{R}(\mathbf{m}_{1}) \cap \mathcal{R}(\mathbf{m}_{2}) = 0$ means that there is no overlapping area between the mask $\mathbf{m}_{1}$ and mask $\mathbf{m}_{2}$. The selected small masks are in the set 
\begin{align}
\Theta  = \{\mathbf{m} \in \mathbf{M}_{small}| \mathcal{R}(\mathbf{m}) \cap \mathcal{R}(\mathbf{M})\ne 0 \},
\label{eq:mask_select}
\end{align}
where $\mathbf{M}_{small}$ means the small mask set that fully covers the image in the next iteration. The optimization strategy $\Omega(\cdot)$ is used in line \ref{line:sliding_space_optimization} of Algorithm \ref{al:Iteration_based_Black-box_Certified_Defense}. It first selects the big mask which satisfies the consistency check (\ie, $\mathcal{CP}=1$) (See yellow mask in Fig.~\ref{fig:optimization}), then chooses the small masks (See blue masks in  Fig.~\ref{fig:optimization}) that intersect with the big mask to be small mask set $\Theta$. It is obvious that the optimization strategy can efficiently reduce the sliding space since the sliding space is reduced from the entire image to a small part.

\subsection{Robustness Certification for Patch Size Estimation Framework}
In this subsection, we give the robustness certification for our iterative black-box certified defense method. Our method contains three main modules: search operation, satisfiability check, and search space reduction. For the search operation module, its robustness is basically guaranteed by the white-box method double-mask algorithm \cite{xiang2022patchcleanser}. Thus we mainly provide robustness certification for the search space reduction and satisfiability check in \textbf{Theorem 1} and \textbf{Theorem 2}, respectively. They give the guarantee to certifiably estimate the size of the patch.\\
\textbf{Theorem 1.} \textit{In search space reduction, among the mask size intervals $[[\eta_{1},\eta_{2}],[\eta_{2},\eta_{3}],\cdots,[\eta_{i},\eta_{i+1}]]$, there must \textbf{exist and only exist} an index $i$, which makes the patch size $v$ locating in an interval $[\eta_{i},\eta_{i+1}]$. The state of $\mathcal{SCP}$ changes from $True$ to $False$ when the search of mask interval changes from $[\eta_{i},\eta_{i+1}]$ to $[\eta_{i+1},\eta_{i+2}]$.}
\begin{align}
    \exists ! i : & v \in [\eta_{i},\eta_{i+1}] (\mathcal{SCP}_s = True) \& \nonumber\\ & v \notin [\eta_{i+1},\eta_{i+2}] (\mathcal{SCP}_s = False).
\label{eq:completness_proofs}
\end{align}
\textit{Existence proof.} Prove \textbf{there exists} a mask size interval $[\eta_{i},\eta_{i+1}]$ that contains the patch size $v$. Considering the initial moment, the mask size is the largest size $\eta_{max}$ and must be bigger than the patch, so the satisfiability check state $\mathcal{SCP}_s=True$. When mask size reduces, there must be an interval $[\eta_{i},\eta_{i+1}]$ containing patch size $v$. Next, when the interval is reduced to $[\eta_{i+1},\eta_{i+2}]$, the mask is smaller than the patch and $\mathcal{SCP}_s = False$.\\
\textit{Uniqueness proof.} Prove there is \textbf{only one} mask size interval $[\eta_{i},\eta_{i+1}]$ contains patch size $v$. Assume there are two different and non-overlap intervals $[\eta_{i},\eta_{i+1}]$ and $[\eta_{j},\eta_{j+1}]$ that both contains patch size $v$. Then the patch size $v$ should simultaneously locate in two non-overlap intervals, which is impossible since $v$ is a fixed value.\\
\textbf{Theorem 2.} \textit{\textbf{If and only if} the mask is larger than the patch, $\exists\mathcal{CP}=1$, as shown in the following formula}
\begin{align}
    (\eta \geq v) \iff (\exists\mathcal{CP}=1).
\label{eq:sat_completness_proofs}
\end{align}
\textit{Proof of Adequacy.} 
When $\eta \geq v$, according to Definition 1 (R-covering) and Definition 2 (Two-mask correctness) in PatchCleanser \cite{xiang2022patchcleanser}, we can infer that in the one-mask prediction, there must have a mask that completely covers the patch and restore the correct prediction. This one-mask image will always output the correct prediction in the second round prediction, and the predictions are consistent. Thus there exists consistency predictions $\mathcal{CP}=1$.\\
\textit{Proof of Necessity.} We use proof by contradiction. The assumed proposition is $\neg ((\exists\mathcal{CP}=1) \rightarrow (\eta \geq v))$. The assumed proposition can be evolved into $(\exists\mathcal{CP}=1) \land (\eta < v)$.
We came to the conclusion that $(\exists\mathcal{CP}=1)$ and $(\eta < v)$ must both be satisfied. However, once the mask size is smaller than the patch size, the image will always be misclassified due to the influence of adversarial patch and there will always come inconsistency predictions $\mathcal{CP}=0$, which fully against $(\exists\mathcal{CP}=1)$. Thus the assumption is overthrown, and the original proposition is established.

\subsection{Algorithm of Patch Size Estimation}
\begin{algorithm}[t]
    \caption{Iterative Patch Size Estimation}
    \label{al:Iteration_based_Black-box_Certified_Defense}
    \KwIn{Image $\mathbf{I}$, Classifier $\mathcal{F}(\cdot)$, Mask set $\mathbf{M}$.}
    \KwOut{Estimated patch size.}
    \BlankLine
    \SetAlgoNoLine
    $\mathbf{U}\leftarrow \phi$  \label{line:prediction_init} \hfill\textcolor{blue}{\Comment{Prediction collection}}
    $y_{prior} \leftarrow{\mathcal{F}(\mathbf{I})}$ \label{line:prior_prediction} \hfill \textcolor{blue}{\Comment{Prior prediction}}
    \For{$\mathbf{m} \in \mathbf{\mathbf{M}[\eta_{max}]}$}
    {\label{line:select_biggest_mask}
    \If{$\mathcal{F}(\mathbf{I}\odot{\mathbf{m}})\not=y_{prior}$}
    {\label{line:basic_judge}
    $\mathbf{U}\leftarrow \mathbf{U}.add({\mathcal{F}(\mathbf{I}\odot{\mathbf{m}})})$ \label{line:prediction_collection}
    }
    }
    \If {$\mathbf{U}=\phi$} 
    {\label{line:judge_clean}
    \textbf{return} $0$ \label{line:return_clean}
    }
    $Iteration \leftarrow 0$  \label{line:iteration_init}\\
    $y_{true} \leftarrow null$ \label{line:correct_label_init}\\
    \For{$\eta \leftarrow{{\eta_{max}}\ \KwTo\ {\eta_{min}}}$} 
    {\label{line:for_loop}
    $\mathbf{\hat{M}} \leftarrow \phi$\\ \label{line:M_hat_init}
    $\mathbf{SCP} \leftarrow \phi$\\   \label{line:SCP_init}
    $\mathbf{L}[Iteration]\leftarrow \eta$\\  \label{line:Mask_size_init}
    \textcolor{blue}{\Comment{One-mask prediction}}
    \For{$\mathbf{m_0} \in {\Omega}(\mathbf{M}[\eta])$}
    {\label{line:sliding_space_optimization}
    $\hat{y} \leftarrow{\mathcal{F}(\mathbf{I}\odot{\mathbf{m_0}}})$ \\
    \If{$\hat{y}\not=y_{prior}$
    \textbf{or} $\hat{y} = y_{true}$}
    {
    $\mathbf{\hat{M}} \leftarrow \mathbf{\hat{M}}.add(\mathbf{m_{0}})$
    \label{line:select_mask_to_set}
    }
    }
    \textcolor{blue}{\Comment{Double-mask prediction}}

    \For{$\mathbf{m_0} \in \mathbf{\hat{M}}$}
    {\label{line:double_mask_prediction}
    $\mathbf{P}    \leftarrow \phi$ \label{line:double_mask_prediction_set}\\ 
    \For{$\mathbf{m_1}\in \mathbf{M}[\eta]$}
    {
    $\mathbf{P}  \leftarrow \mathbf{P}.add({\mathcal{F}(\mathbf{I}\odot{\mathbf{m_0}}\odot{\mathbf{m_1}})})$ \label{line:double_mask_prediction_collection}\\
    }
    $\mathcal{CP},y_{con} \leftarrow \mathrm{ConsistencyCheck(\mathbf{P})}$ \label{line:consistency_check}\\
    $\mathbf{SCP} \leftarrow \mathbf{SCP}.add(\mathcal{CP})$ \label{line:put_consistency_result_set}\\
    \If{$Iteration = 0$ \textbf{and} \ $\mathcal{CP}=1$}
    {\label{line:first_iteration_label_judge}
    $y_{true} \leftarrow{y_{con}}$ \label{line:first_iteration_label_set}
    }
    \If{ $y_{con} \not= y_{true}$ \textbf{and} \ $\mathcal{CP}=1$}
    {\label{line:y_true_compare}
    \textbf{return $\mathbf{L}[Iteration-1]$} \label{line:output_mask_size_1}
    }
    }
    \If {$({\forall}\mathcal{CP}\in \mathbf{SCP}, \mathcal{CP}=0)$}
    {\label{line:output_judge}
    \textbf{return $\mathbf{L}[Iteration-1]$}  \label{line:output_mask_size_2}
    }
    $Iteration \leftarrow Iteration + 1$\\
    }
\textbf{return $\mathbf{L}[Iteration]$}
\end{algorithm}
In Algorithm \ref{al:Iteration_based_Black-box_Certified_Defense}, we specifically show how to estimate the patch size. Please note that the procedure is only a little different from the description of the method because we aim to show the algorithm more clearly by dividing it into two parts (actually the two parts can be integrated into a complete iteration as introduced in the former sections). The first part describes the judgment on whether the image is clean or adversarial. The second part is only designed for adversarial images. The input is an image $\mathbf{I}$, a classifier $\mathcal{F}(\cdot)$, and a multi-scale mask set $\mathbf{M}$. The output is the estimated patch size.  

The algorithm first judges whether the image is clean or being attacked in lines 1-7. In line \ref{line:prediction_init}, initialize the consistency prediction collection. In line \ref{line:prior_prediction}, calculate the $y_{prior}$ as the prior of the input image. In line \ref{line:select_biggest_mask}, select the mask of the biggest mask size. Here select many same-size masks which have different positions in \textit{for} loop. In lines \ref{line:basic_judge}-\ref{line:prediction_collection}, judge whether the one-masked images have the same prediction as $y_{prior}$ and put the predictions into the set $\mathbf{U}$. In lines \ref{line:judge_clean}-\ref{line:return_clean}, we can confirm that the input image is clean if $\mathbf{U}$ is an empty set and return the 0 as patch size.

Lines 8-33 describe the iteration on how to estimate the patch size for adversarial images. In line \ref{line:iteration_init}-\ref{line:correct_label_init}, we initialize a variable $Iteration$ to count the number of iterations and initialize $y_{true}$ to represent the ground truth label of the adversarial image. In line \ref{line:for_loop}, we start the iteration with the maximum mask size and decrease it round by round. In lines \ref{line:M_hat_init}-\ref{line:Mask_size_init}, we initialize $\mathbf{\hat{M}}$ to record the selected masks in one-mask prediction, initialize $\mathbf{SCP}$ to record the inconsistency check results (\ie, $\mathcal{CP}$), initialize $\mathbf{L}$ to record the mask size in each iteration. In line \ref{line:sliding_space_optimization}-\ref{line:select_mask_to_set}, we first do one mask prediction on the image and obtain the classification result. The masks that lead to $\hat{y}\not=y_{prior}$ or $\hat{y} = y_{true}$ are selected into the $\mathbf{\hat{M}}$. In line \ref{line:double_mask_prediction}, perform \textit{for} loop with each one-mask as a group. In lines \ref{line:double_mask_prediction_set}-\ref{line:double_mask_prediction_collection}, collecting the prediction results of double-masking. In line \ref{line:consistency_check}, do a consistency check on $\mathbf{P}$ to see whether the images in the group have the same classification results. The function $\mathrm{ConsistencyCheck(\cdot)}$ returns $\mathcal{CP}$ and $y_{con}$, where $\mathcal{CP}$ is the consistency result (1 if same, 0 if different), $y_{con}$ is the consistent label if $\mathcal{CP}=1$. In line \ref{line:put_consistency_result_set}, we put the consistency results in a set $\mathbf{SCP}$. In line \ref{line:first_iteration_label_judge}-\ref{line:first_iteration_label_set}, it confirms the ground truth label of the unpatched adversarial image (\ie, the clean image without the patch). The label is confirmed in the first iteration. In line \ref{line:y_true_compare}-\ref{line:output_mask_size_1}, the label returned by the consistency result is not the same as the ground truth label, which means the mask is smaller than the patch and the algorithm should return the mask size of the previous iteration. In line \ref{line:output_judge}-\ref{line:output_mask_size_2}, when all the inconsistency checks are not the same, the mask is smaller than the patch and the algorithm should return the mask size of the previous iteration.

\section{Experiment}
IBCD is a two-stage black-box certified defense method and our work mainly focuses on the first stage (\ie, patch size estimation). Thus we conduct three perspectives to demonstrate the search efficiency and estimation accuracy. First, we compare the accuracy calculated under the black-box and the white-box settings to show the deviation caused by the gap between the exact patch size and the estimated patch size. Second, we compare the proposed search algorithm with brute-force search to show the efficiency of our method. Third, we design a comparative experiment to show the effect of the proposed sliding space optimization strategy. 

\subsection{Experiment Setup}
\noindent\textbf{Datasets.} We carry out experiments on two datasets CIFAR10 and ImageNet respectively. ImageNet \cite{5206848} has 1,000 classes of different categories. We randomly choose 100,000 images with the size of 224 $\times$ 224 in its test set. CIFAR10 \cite{krizhevsky2009learning} is a famous classification dataset. We choose all the data in the test set (\ie, 10,000 images of size 32 $\times$ 32). 

\noindent\textbf{Target models.} We choose a classic network (\ie, ResNet50 \cite{he2016deep}) and a state-of-the-art network (\ie, ViT \cite{dosovitskiy2021an}) for the experiment. They are popular and representative.

\noindent\textbf{Adversarial patch.}
To verify the effectiveness of IBCD, we assume the adversarial patch has the \textit{strongest attack ability} in theory (\ie, the patch will always mislead the model once it has pixels exposed in the image). The patch is of side length $v$ and can be layout in arbitrary positions.

\noindent\textbf{Metrics.} To evaluate the performance of IBCD, we compare it with the white-box certified defense to show the error range. The metrics of certified defense are \textit{clean accuracy} and \textit{certified accuracy}. The \textit{clean accuracy} is the fraction of clean test images that can be correctly classified by the defended model. The \textit{certified accuracy} is the fraction of test images that the classification is correct under certain patch attacks. In addition, we also use the \textit{pre-example searches} to analyze the search efficiency of our black-box condition.

\noindent\textbf{Implementation details.}
In the ideal case, the sliding stride should be $s=1$, which can guarantee that the mask set can fully cover the patch in any position of the image if the patch is smaller than the mask. However, only a few images can pass the certification, which leads to the lack of input images for our patch size estimation method. Since our contribution mainly focuses on patch size estimation, thus we empirically and restrained to magnify $s$ to an acceptable value to achieve more certified images for fully demonstrating the performance of our patch size estimation method. For CIFAR10, we set sliding stride $s=5$ and the reduction interval of the multi-scale mask set $\mathbf{M}$ to be 2. For ImageNet, we set sliding stride $s=40$ and the reduction interval of the multi-scale mask set $\mathbf{M}$ to be 20. The patch size used to attack CIFAR10 is in the range $[2,16]$ and that used to attack ImageNet is in the range $[2,112]$. All the experiments were run on the ubuntu 18.04 system, Pytorch 1.7.0, and CUDA 11.0 with an AMD EPYC 7543 32-Core Processor CPU, 80GB of RAM, and an NVIDIA GeForce RTX 3090 GPU of 24GB RAM.

\subsection{Accuracy Fluctuation between Black-box and White-box Certified Defense}
In this experiment, we aim to evaluate the accuracy of the estimated patch size and compare the accuracy (certified \& clean) deviation between the black-box and white-box settings. 
\begin{table*}[]
\caption{\textbf{Accuracy fluctuation between black-box and white-box certified defense.} For each patch size, we calculate the estimated size and corresponding accuracy. The fluctuation rates of the accuracy are shown in the last column.}
\resizebox{0.9\linewidth}{!}{%
\begin{tabular}{l|ccc|ccc|cc}
\toprule 
\multirow{1}{*}{} & actual size & certified acc (\%)& clean acc (\%)& estimated size & certified acc (\%)& clean acc (\%)& \makecell[c]{certified acc \\ fluctuation rate}  (\%)& \makecell[c]{clean acc \\ fluctuation rate}  (\%)\tabularnewline
\midrule 
\multirow{5}{*}{CIFAR10-ViT} & 15 & 10.00 & 10.00 & 18.61 (19) & 4.73 & 10.08 & 52.70 & 0.80\tabularnewline
 & 12 & 6.73 & 10.79 & 16.40 (17) & 10.00 & 10.00 & 48.59 & 7.32\tabularnewline
 & 9 & 7.97 & 10.87 & 13.86 (14) & 7.54 & 9.77 & 5.40 & 10.12\tabularnewline
 & 6 & 9.81 & 10.03 & 11.09 (12) & 6.73 & 10.79 & 31.40 & 7.58\tabularnewline
 & 3 & 11.91 & 13.35 & 8.56 (9) & 7.97 & 10.87 & 33.08 & 18.58\tabularnewline
\midrule 
\multirow{5}{*}{CIFAR10-ResNet50} & 15 & 0.10 & 45.72 & 19.32 (20) & 0.05 & 23.87 & 50.00 & 47.79\tabularnewline
 & 12 & 0.29 & 65.25 & 16.44 (17) & 0.09 & 36.88 & 68.97 & 43.48\tabularnewline
 & 9 & 0.77 & 77.59 & 13.47 (14) & 0.10 & 50.12 & 87.01 & 35.40\tabularnewline
 & 6 & 3.66 & 84.92 & 10.37 (11) & 0.21 & 68.66 & 94.26 & 19.15\tabularnewline
 & 3 & 20.62 & 87.77 & 8.45 (9) & 0.77 & 77.59 & 96.27 & 11.60\tabularnewline
\midrule 
\multirow{5}{*}{ImageNet-ViT} & 110 & 21.95 & 81.17 & 141 & 9.63 & 77.54 & 56.13 & 4.47\tabularnewline
 & 90 & 31.69 & 82.61 & 131 & 15.61 & 79.88 & 50.74 & 3.30\tabularnewline
 & 70 & 43.35 & 84.96 & 107 & 22.93 & 81.72 & 47.10 & 3.81\tabularnewline
 & 50 & 53.21 & 84.15 & 85 & 31.85 & 81.31 & 40.14 & 3.37\tabularnewline
 & 30 & 57.92 & 85.90 & 51 & 53.29 & 84.15 & 7.99 & 2.04\tabularnewline
\bottomrule 
\end{tabular}
}
\label{tab:Accuracy_fluctuation_between_black-box_and_white_box_certified_defense}
\end{table*}
We show the estimation results in Table \ref{tab:Accuracy_fluctuation_between_black-box_and_white_box_certified_defense}. The target datasets and models are listed in the first column. The certified accuracy and clean accuracy are calculated by PatchCleanser with the given actual or estimated patch size. The average estimated patch size is usually in float format and the size should be an integer, thus we take the upper integer bound to guarantee the certified robustness. The second column shows the actual patch size of the adversarial patch, the corresponding certified accuracy (certified acc), and clean accuracy (clean acc). The third column shows the estimated patch size of the adversarial patch, the corresponding certified accuracy (certified acc), and clean accuracy (clean acc). The fourth column shows the fluctuation rate between the accuracy calculated by actual patch size and estimated patch size. The formula for calculating the fluctuation rate is
\begin{align}
Acc_{flu}=\left | \frac{Acc_{white}-Acc_{black}}{Acc_{white}}  \right |,
\label{eq:accuracy_calculate_formula}
\end{align}
where $Acc_{white}$ and $Acc_{black}$ mean the accuracy (certified and clean) calculated with actual and estimated patch size respectively. A smaller $Acc_{flu}$ is better.
We use different patch sizes to attack the target model (\ie, ResNet50 or ViT), which can give a comprehensive result on the performance of patch size estimation. For each patch size, we use Algorithm \ref{al:Iteration_based_Black-box_Certified_Defense} to obtain the estimated patch size. The estimated patch sizes are always bigger than the actual counterpart, which provides guarantees of fully covering the patch.

For the CIFAR10-ViT, we can find that the fluctuation rate of certified accuracy does not exceed 52.70\%, and the clean accuracy does not exceed 18.58\%. For CIFAR10-ResNet50, the fluctuation rate is worse than that in CIFAR10-ViT, which means ViT is a better model for double masking-based certified defense. For ImageNet-ViT, the fluctuation rate of certified accuracy does not exceed 56.13\%, and the clean accuracy does not exceed 4.47\%. In summary, the accuracy fluctuation between the black-box method and the white-box method is within an acceptable range. 
In addition, there is an obvious phenomenon that when the actual patch size is large, the certified accuracy is very low (\eg, size=15 is nearly half the length of the CIFAR10 images). This is the defect of the current certified patch defense method, it is often at a loss when faced with a large-size patch. However, under such extreme conditions, the fluctuation of our proposed black-box method is not more than half the accuracy of the white-box method. 


\subsection{Search Efficiency}
We aim to show the search efficiency of the proposed patch size estimation method, which is an important metric to determine the practicality of the method in the physical world. 
With the fixed size and the position of the adversarial patch, we set the sliding step of the mask to be $s = 7$ and only change the interval of mask reduction. As shown in Table \ref{tab:search_efficiency}, the first column represents the interval of mask reduction on the multi-scale mask set $\mathbf{M}$. In the first row, we record the average search number and the average search time to evaluate the patch size estimation process. When the interval of mask reduction is 1, it means brute-force search, which leads to the most search number (\eg, 206) and searches time (\eg, 0.92s and 1.41s). Obviously, with the increment of mask reduction interval, the search number and the search time are significantly reduced. In addition, with the same mask reduction interval, ViT and ResNet50 have the same search number, which also shows the architecture-agnostic property of IBCD.
\begin{table}
\caption{Comparison of search efficiency between our patch size estimation method and brute-force search.}
\resizebox{0.9\linewidth}{!}{
\begin{tabular}{c|cccc}
\toprule 
\multirow{2}{*}{\makecell[c]{Reduction \\ interval}} & \multicolumn{2}{c}{CIFAR10-ViT} & \multicolumn{2}{c}{CIFAR10-ResNet50}\tabularnewline
 & search num (\#) & time (s) & search num (\#) & time (s)\tabularnewline
\midrule 
\textbf{1} & \textbf{206} & \textbf{0.92} & \textbf{206} & \textbf{1.41}\tabularnewline
2 & 104 & 0.77 & 104 & 0.79\tabularnewline
3 & 100 & 0.83 & 100 & 0.80\tabularnewline
4 & 68 & 0.69 & 68 & 0.73\tabularnewline
5 & 60 & 0.65 & 60 & 0.56\tabularnewline
6 & 44 & 0.49 & 44 & 0.33\tabularnewline
7 & 42 & 0.43 & 42 & 0.17\tabularnewline
\bottomrule 
\end{tabular}
}
\label{tab:search_efficiency}
\end{table}
\subsection{Evaluation on Optimization Strategy}
In this subsection, we design a comparative experiment to show the effectiveness of the sliding space optimization strategy. We calculate the search efficiency on 1,000 robust certified images and show the result in Table \ref{tab:Evaluation_optimization_strategy}. In the first column, we list different patch sizes to comprehensively evaluate the search efficiency. In the first row, we use ``CIFAR10-Vanilla'' and ``CIFAR10-SlidingOpt'' to represent the search method without/with sliding optimization strategy respectively. In each cell, we demonstrate the search number. It is obvious that after applying the sliding space optimization strategy, the search number significantly decreases to about half. 
\begin{table}
\centering
\caption{Search efficiency w/wo sliding space optimization.}
\resizebox{0.8\linewidth}{!}{
\begin{tabular}{c|cc}
\toprule 
\multirow{2}{*}{Patch size} & \multirow{1}{*}{CIFAR10-Vanilla} & CIFAR10-SlidingOpt\tabularnewline
 & search num (\#) & search num (\#)\tabularnewline
\midrule 
11 & 175 & 96\tabularnewline
7 & 286 & 138\tabularnewline
4 & 322 & 170\tabularnewline
\bottomrule 
\end{tabular}
}
\label{tab:Evaluation_optimization_strategy}
\end{table}



\subsection{Discussion}
\noindent\textbf{Limitation of white-box certified defense on certified accuracy.} 
At present, in the field of adversarial patch certified defense, the certified accuracy of existing white-box methods is still low and their performance represents the upper bound of our black-box certified defense method. Thus the low certified accuracy in Table \ref{tab:Accuracy_fluctuation_between_black-box_and_white_box_certified_defense} is due to the limitation of the white-box method, not caused by our method. In the future, with the improvement of white-box certified defense methods, our method can also achieve better results.

\section{Conclusion}
In this paper, we retrospect the architecture-agnostic certified defense methods and find that they have an obvious defect. That is, they inevitably need to access the patch size of the adversarial patch attack, which is impractical and unreasonable in the real world. To address this problem, we propose a two-stage black-box certified defense. The method estimates the patch size in the first stage and outputs the accuracy in the second stage. We believe this method provides a choice for further applying certified defense into the physical world. In future work, we aim to simplify the method into a one-stage end-to-end method for better efficiency. 

\bibliographystyle{ACM-Reference-Format}
\bibliography{sample-base}


\begin{thebibliography}{20}


\ifx \showCODEN    \undefined \def \showCODEN     #1{\unskip}     \fi
\ifx \showDOI      \undefined \def \showDOI       #1{#1}\fi
\ifx \showISBNx    \undefined \def \showISBNx     #1{\unskip}     \fi
\ifx \showISBNxiii \undefined \def \showISBNxiii  #1{\unskip}     \fi
\ifx \showISSN     \undefined \def \showISSN      #1{\unskip}     \fi
\ifx \showLCCN     \undefined \def \showLCCN      #1{\unskip}     \fi
\ifx \shownote     \undefined \def \shownote      #1{#1}          \fi
\ifx \showarticletitle \undefined \def \showarticletitle #1{#1}   \fi
\ifx \showURL      \undefined \def \showURL       {\relax}        \fi
\providecommand\bibfield[2]{#2}
\providecommand\bibinfo[2]{#2}
\providecommand\natexlab[1]{#1}
\providecommand\showeprint[2][]{arXiv:#2}

\bibitem[Brown et~al\mbox{.}(2017)]%
        {brown2017adversarial}
\bibfield{author}{\bibinfo{person}{Tom~B Brown}, \bibinfo{person}{Dandelion
  Man{\'e}}, \bibinfo{person}{Aurko Roy}, \bibinfo{person}{Mart{\'\i}n Abadi},
  {and} \bibinfo{person}{Justin Gilmer}.} \bibinfo{year}{2017}\natexlab{}.
\newblock \showarticletitle{Adversarial patch}.
\newblock \bibinfo{journal}{\emph{arXiv preprint arXiv:1712.09665}}
  (\bibinfo{year}{2017}).
\newblock


\bibitem[Chen et~al\mbox{.}(2022)]%
        {chen2022towards}
\bibfield{author}{\bibinfo{person}{Zhaoyu Chen}, \bibinfo{person}{Bo Li},
  \bibinfo{person}{Jianghe Xu}, \bibinfo{person}{Shuang Wu},
  \bibinfo{person}{Shouhong Ding}, {and} \bibinfo{person}{Wenqiang Zhang}.}
  \bibinfo{year}{2022}\natexlab{}.
\newblock \showarticletitle{Towards Practical Certifiable Patch Defense with
  Vision Transformer}. In \bibinfo{booktitle}{\emph{Proceedings of the IEEE/CVF
  Conference on Computer Vision and Pattern Recognition}}.
  \bibinfo{pages}{15148--15158}.
\newblock


\bibitem[Chiang et~al\mbox{.}(2020)]%
        {chiang2020certified}
\bibfield{author}{\bibinfo{person}{Ping-yeh Chiang}, \bibinfo{person}{Renkun
  Ni}, \bibinfo{person}{Ahmed Abdelkader}, \bibinfo{person}{Chen Zhu},
  \bibinfo{person}{Christoph Studer}, {and} \bibinfo{person}{Tom Goldstein}.}
  \bibinfo{year}{2020}\natexlab{}.
\newblock \showarticletitle{Certified defenses for adversarial patches}.
\newblock \bibinfo{journal}{\emph{arXiv preprint arXiv:2003.06693}}
  (\bibinfo{year}{2020}).
\newblock


\bibitem[Deng et~al\mbox{.}(2009)]%
        {5206848}
\bibfield{author}{\bibinfo{person}{Jia Deng}, \bibinfo{person}{Wei Dong},
  \bibinfo{person}{Richard Socher}, \bibinfo{person}{Li-Jia Li},
  \bibinfo{person}{Kai Li}, {and} \bibinfo{person}{Li Fei-Fei}.}
  \bibinfo{year}{2009}\natexlab{}.
\newblock \showarticletitle{ImageNet: A large-scale hierarchical image
  database}. In \bibinfo{booktitle}{\emph{2009 IEEE Conference on Computer
  Vision and Pattern Recognition}}. \bibinfo{pages}{248--255}.
\newblock
\urldef\tempurl%
\url{https://doi.org/10.1109/CVPR.2009.5206848}
\showDOI{\tempurl}


\bibitem[Dosovitskiy et~al\mbox{.}(2021)]%
        {dosovitskiy2021an}
\bibfield{author}{\bibinfo{person}{Alexey Dosovitskiy}, \bibinfo{person}{Lucas
  Beyer}, \bibinfo{person}{Alexander Kolesnikov}, \bibinfo{person}{Dirk
  Weissenborn}, \bibinfo{person}{Xiaohua Zhai}, \bibinfo{person}{Thomas
  Unterthiner}, \bibinfo{person}{Mostafa Dehghani}, \bibinfo{person}{Matthias
  Minderer}, \bibinfo{person}{Georg Heigold}, \bibinfo{person}{Sylvain Gelly},
  \bibinfo{person}{Jakob Uszkoreit}, {and} \bibinfo{person}{Neil Houlsby}.}
  \bibinfo{year}{2021}\natexlab{}.
\newblock \showarticletitle{An Image is Worth 16x16 Words: Transformers for
  Image Recognition at Scale}. In \bibinfo{booktitle}{\emph{International
  Conference on Learning Representations}}.
\newblock
\urldef\tempurl%
\url{https://openreview.net/forum?id=YicbFdNTTy}
\showURL{%
\tempurl}


\bibitem[Hayes(2018)]%
        {hayes2018visible}
\bibfield{author}{\bibinfo{person}{Jamie Hayes}.}
  \bibinfo{year}{2018}\natexlab{}.
\newblock \showarticletitle{On visible adversarial perturbations \& digital
  watermarking}. In \bibinfo{booktitle}{\emph{Proceedings of the IEEE
  Conference on Computer Vision and Pattern Recognition Workshops}}.
  \bibinfo{pages}{1597--1604}.
\newblock


\bibitem[He et~al\mbox{.}(2016)]%
        {he2016deep}
\bibfield{author}{\bibinfo{person}{Kaiming He}, \bibinfo{person}{Xiangyu
  Zhang}, \bibinfo{person}{Shaoqing Ren}, {and} \bibinfo{person}{Jian Sun}.}
  \bibinfo{year}{2016}\natexlab{}.
\newblock \showarticletitle{Deep residual learning for image recognition}. In
  \bibinfo{booktitle}{\emph{Proceedings of the IEEE conference on computer
  vision and pattern recognition}}. \bibinfo{pages}{770--778}.
\newblock


\bibitem[Karmon et~al\mbox{.}(2018)]%
        {karmon2018lavan}
\bibfield{author}{\bibinfo{person}{Danny Karmon}, \bibinfo{person}{Daniel
  Zoran}, {and} \bibinfo{person}{Yoav Goldberg}.}
  \bibinfo{year}{2018}\natexlab{}.
\newblock \showarticletitle{Lavan: Localized and visible adversarial noise}. In
  \bibinfo{booktitle}{\emph{International Conference on Machine Learning}}.
  PMLR, \bibinfo{pages}{2507--2515}.
\newblock


\bibitem[Krizhevsky et~al\mbox{.}(2009)]%
        {krizhevsky2009learning}
\bibfield{author}{\bibinfo{person}{Alex Krizhevsky}, \bibinfo{person}{Geoffrey
  Hinton}, {et~al\mbox{.}}} \bibinfo{year}{2009}\natexlab{}.
\newblock \showarticletitle{Learning multiple layers of features from tiny
  images}.
\newblock  (\bibinfo{year}{2009}).
\newblock


\bibitem[Levine and Feizi(2020)]%
        {levine2020randomized}
\bibfield{author}{\bibinfo{person}{Alexander Levine} {and}
  \bibinfo{person}{Soheil Feizi}.} \bibinfo{year}{2020}\natexlab{}.
\newblock \showarticletitle{(De) Randomized smoothing for certifiable defense
  against patch attacks}.
\newblock \bibinfo{journal}{\emph{Advances in Neural Information Processing
  Systems}}  \bibinfo{volume}{33} (\bibinfo{year}{2020}),
  \bibinfo{pages}{6465--6475}.
\newblock


\bibitem[Lin et~al\mbox{.}(2020)]%
        {lin2020certified}
\bibfield{author}{\bibinfo{person}{Wan-Yi Lin}, \bibinfo{person}{Fatemeh
  Sheikholeslami}, \bibinfo{person}{Leslie Rice}, \bibinfo{person}{J~Zico
  Kolter}, {et~al\mbox{.}}} \bibinfo{year}{2020}\natexlab{}.
\newblock \showarticletitle{Certified robustness against physically-realizable
  patch attack via randomized cropping}.
\newblock  (\bibinfo{year}{2020}).
\newblock


\bibitem[Liu et~al\mbox{.}(2018)]%
        {liu2018dpatch}
\bibfield{author}{\bibinfo{person}{Xin Liu}, \bibinfo{person}{Huanrui Yang},
  \bibinfo{person}{Ziwei Liu}, \bibinfo{person}{Linghao Song},
  \bibinfo{person}{Hai Li}, {and} \bibinfo{person}{Yiran Chen}.}
  \bibinfo{year}{2018}\natexlab{}.
\newblock \showarticletitle{Dpatch: An adversarial patch attack on object
  detectors}.
\newblock \bibinfo{journal}{\emph{arXiv preprint arXiv:1806.02299}}
  (\bibinfo{year}{2018}).
\newblock


\bibitem[Metzen and Yatsura(2021)]%
        {metzen2021efficient}
\bibfield{author}{\bibinfo{person}{Jan~Hendrik Metzen} {and}
  \bibinfo{person}{Maksym Yatsura}.} \bibinfo{year}{2021}\natexlab{}.
\newblock \showarticletitle{Efficient certified defenses against patch attacks
  on image classifiers}.
\newblock \bibinfo{journal}{\emph{arXiv preprint arXiv:2102.04154}}
  (\bibinfo{year}{2021}).
\newblock


\bibitem[Naseer et~al\mbox{.}(2019)]%
        {naseer2019local}
\bibfield{author}{\bibinfo{person}{Muzammal Naseer}, \bibinfo{person}{Salman
  Khan}, {and} \bibinfo{person}{Fatih Porikli}.}
  \bibinfo{year}{2019}\natexlab{}.
\newblock \showarticletitle{Local gradients smoothing: Defense against
  localized adversarial attacks}. In \bibinfo{booktitle}{\emph{2019 IEEE Winter
  Conference on Applications of Computer Vision (WACV)}}. IEEE,
  \bibinfo{pages}{1300--1307}.
\newblock


\bibitem[Salman et~al\mbox{.}(2022)]%
        {salman2022certified}
\bibfield{author}{\bibinfo{person}{Hadi Salman}, \bibinfo{person}{Saachi Jain},
  \bibinfo{person}{Eric Wong}, {and} \bibinfo{person}{Aleksander Madry}.}
  \bibinfo{year}{2022}\natexlab{}.
\newblock \showarticletitle{Certified patch robustness via smoothed vision
  transformers}. In \bibinfo{booktitle}{\emph{Proceedings of the IEEE/CVF
  Conference on Computer Vision and Pattern Recognition}}.
  \bibinfo{pages}{15137--15147}.
\newblock


\bibitem[Tramer et~al\mbox{.}(2020)]%
        {tramer2020adaptive}
\bibfield{author}{\bibinfo{person}{Florian Tramer}, \bibinfo{person}{Nicholas
  Carlini}, \bibinfo{person}{Wieland Brendel}, {and}
  \bibinfo{person}{Aleksander Madry}.} \bibinfo{year}{2020}\natexlab{}.
\newblock \showarticletitle{On adaptive attacks to adversarial example
  defenses}.
\newblock \bibinfo{journal}{\emph{Advances in Neural Information Processing
  Systems}}  \bibinfo{volume}{33} (\bibinfo{year}{2020}),
  \bibinfo{pages}{1633--1645}.
\newblock


\bibitem[Xiang et~al\mbox{.}(2021)]%
        {xiang2021patchguard}
\bibfield{author}{\bibinfo{person}{Chong Xiang}, \bibinfo{person}{Arjun~Nitin
  Bhagoji}, \bibinfo{person}{Vikash Sehwag}, {and} \bibinfo{person}{Prateek
  Mittal}.} \bibinfo{year}{2021}\natexlab{}.
\newblock \showarticletitle{$\{$PatchGuard$\}$: A Provably Robust Defense
  against Adversarial Patches via Small Receptive Fields and Masking}. In
  \bibinfo{booktitle}{\emph{30th USENIX Security Symposium (USENIX Security
  21)}}. \bibinfo{pages}{2237--2254}.
\newblock


\bibitem[Xiang et~al\mbox{.}(2022)]%
        {xiang2022patchcleanser}
\bibfield{author}{\bibinfo{person}{Chong Xiang}, \bibinfo{person}{Saeed
  Mahloujifar}, {and} \bibinfo{person}{Prateek Mittal}.}
  \bibinfo{year}{2022}\natexlab{}.
\newblock \showarticletitle{$\{$PatchCleanser$\}$: Certifiably Robust Defense
  against Adversarial Patches for Any Image Classifier}. In
  \bibinfo{booktitle}{\emph{31st USENIX Security Symposium (USENIX Security
  22)}}. \bibinfo{pages}{2065--2082}.
\newblock


\bibitem[Xu et~al\mbox{.}(2020)]%
        {xu2020adversarial}
\bibfield{author}{\bibinfo{person}{Kaidi Xu}, \bibinfo{person}{Gaoyuan Zhang},
  \bibinfo{person}{Sijia Liu}, \bibinfo{person}{Quanfu Fan},
  \bibinfo{person}{Mengshu Sun}, \bibinfo{person}{Hongge Chen},
  \bibinfo{person}{Pin-Yu Chen}, \bibinfo{person}{Yanzhi Wang}, {and}
  \bibinfo{person}{Xue Lin}.} \bibinfo{year}{2020}\natexlab{}.
\newblock \showarticletitle{Adversarial t-shirt! evading person detectors in a
  physical world}. In \bibinfo{booktitle}{\emph{Computer Vision--ECCV 2020:
  16th European Conference, Glasgow, UK, August 23--28, 2020, Proceedings, Part
  V 16}}. Springer, \bibinfo{pages}{665--681}.
\newblock


\bibitem[Zhang et~al\mbox{.}(2020)]%
        {zhang2020clipped}
\bibfield{author}{\bibinfo{person}{Zhanyuan Zhang}, \bibinfo{person}{Benson
  Yuan}, \bibinfo{person}{Michael McCoyd}, {and} \bibinfo{person}{David
  Wagner}.} \bibinfo{year}{2020}\natexlab{}.
\newblock \showarticletitle{Clipped bagnet: Defending against sticker attacks
  with clipped bag-of-features}. In \bibinfo{booktitle}{\emph{2020 IEEE
  Security and Privacy Workshops (SPW)}}. IEEE, \bibinfo{pages}{55--61}.
\newblock


\end{thebibliography}

\end{document}